\documentclass[runningheads]{llncs}

\usepackage{iciap}

\usepackage{colortbl} 
\usepackage{booktabs}        
\usepackage{arydshln}        
\usepackage{wrapfig}
\newcommand{\AAAA}{PA-ATM\xspace}
\newcommand{\BBBB}{PH-ATM\xspace}
\usepackage{iciapabbrv}

\usepackage{graphicx}
\usepackage{booktabs}
\usepackage{multirow} 

\definecolor{forestGreen}{RGB}{34, 139, 34}
\definecolor{firebrick}{RGB}{178, 34, 34}

\usepackage[accsupp]{axessibility}  

\usepackage[pagebackref,breaklinks,colorlinks,citecolor=iciapblue]{hyperref}

\usepackage{orcidlink}
\usepackage{natbib}

\definecolor{customRed}{RGB}{237,110,139}
\definecolor{customGreen}{RGB}{101,205,172}
\definecolor{customYellow}{RGB}{255,218,129}

\begin{document}

\title{ATM: Improving Model Merging by Alternating Tuning and Merging} 

\titlerunning{Abbreviated paper title}

\author{Luca Zhou* \and
Daniele Solombrino* \and
Donato Crisostomi \and
Maria Sofia Bucarelli \and
Fabrizio Silvestri \and
Emanuele Rodolà}

\authorrunning{L.Zhou et al.}

\institute{
Sapienza University of Rome, Rome, Italy \\
\email{zhou.2135393@studenti.uniroma1.it \\\{solombrino, crisostomi, rodola\}@di.uniroma1.it}
\\
\email{\{bucarelli, fsilvestri\}@diag.uniroma1.it}
}

\maketitle

\begin{abstract}
  Model merging has emerged as a cost-efficient approximation to multitask learning. Among merging strategies, task arithmetic \citep{task-vectors} is notable for its simplicity and effectiveness. In this work, we provide a theoretical motivation for task vectors by highlighting that, under single-epoch full-batch gradient descent, they are equivalent to multitask gradients. This insight leads us to reinterpret model merging as a single step in an iterative procedure that \textbf{A}lternates between \textbf{T}uning and \textbf{M}erging (ATM). We propose two applications of ATM: (1) as an alternative to multitask learning in scenarios where data sharing is restricted (e.g., federated settings), and (2) as a lightweight refinement step to improve existing model merging methods using a small validation set. Experiments across diverse vision tasks demonstrate the effectiveness of ATM.
  
\end{abstract}

\section{Introduction}
The standard pretrain-and-finetune paradigm becomes storage-intensive for multiple tasks, as it requires a separate model for each. Model merging mitigates this by combining task-specific models into a single network. Among these methods, \textit{task arithmetic} \citep{task-vectors} is prominent for its simplicity and effectiveness. It operates on a {\em task vector}, $\tau_{i} = \theta_{i} - \theta_0$, which represents the difference between a model $\theta_i$ finetuned on task $t_i$ and the pretrained model $\theta_0$. To create a multitask model, task arithmetic adds a scaled sum of these task vectors, controlled by a coefficient $\alpha$, back to the pretrained weights.

\begin{figure}
    \centering
    \vspace{-0.5cm}
    \begin{tikzpicture}
        \node[anchor=south west, inner sep=0] (image) at (0,0) {\includegraphics[width=1\linewidth, bb=0 0 2845 1042]{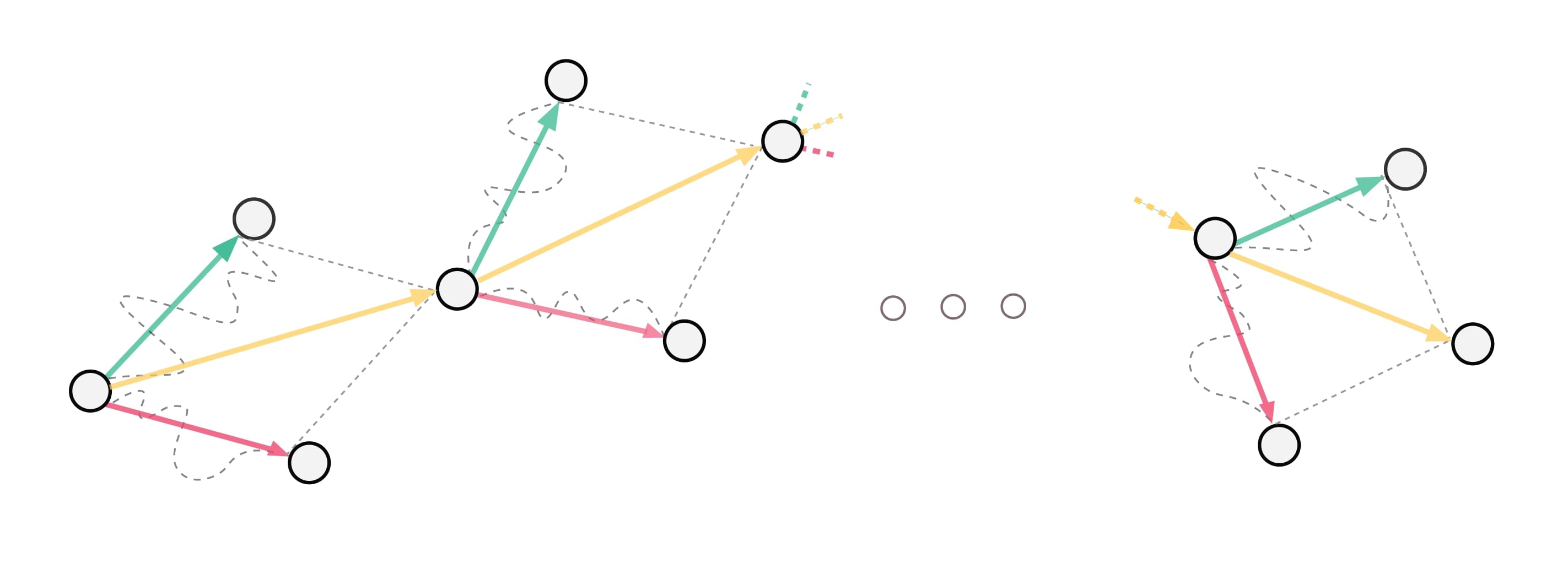}};

        \node at (0.45, 1.82){\tiny $\theta_{\text{base}}^{(0)}$};

        \node at (3.27, 2.6){\tiny $\theta_{\text{base}}^{(1)}$};

        \node at (5.8, 3.75){\tiny $\theta_{\text{base}}^{(2)}$};

        \node at (9.25, 3.03) {\tiny $\theta_{\text{base}}^{(h-1)}$};
        
        \node at (11.15, 2.2){\tiny $\theta_{\text{base}}^{(h)}$};



    \end{tikzpicture}
    \vspace{-1.5em}
    \caption{The ATM framework, illustrated up to iteration $h$ with two tasks (\textcolor{customRed}{\textit{red}} and \textcolor{customGreen}{\textit{green}}). In each iteration, the \textbf{\textit{Tune}} step finetunes the pretrained model $\theta_{\text{base}}^{(0)}$ separately on both tasks, and the \textbf{\textit{Merge}} step aggregates the task vectors and applies the resulting multitask vector (\textcolor{customYellow}{\textit{yellow}}) to the base model. This process repeats, with each iteration using the updated model as the new base, continuing until a stopping condition is met.
    }
    \label{fig:ATM diagram}
\end{figure}

In this paper, we explain the effectiveness of task arithmetic by linking task vectors to the gradients of the task losses. We start by noticing that when a model is finetuned for a single epoch using gradient descent (GD), the corresponding task vector is a scaled additive inverse of the loss gradient. Similarly, the multitask vector, obtained by summing individual task vectors, is equivalent to the additive inverse of the {\em average} of task-specific loss gradients. Thus, task addition is analogous to performing a GD step on the sum of the average task losses.

This perspective recasts task arithmetic as a single, large, and noisy gradient descent (GD) step toward a joint-task objective. A key implication is that this one-step process risks overshooting the multitask optimum. Within this framework, the scaling factor, typically optimized on a validation set \citep{task-vectors}, functions as the effective learning rate for this single update.

Building on this insight, we introduce \textbf{A}lternating \textbf{T}uning and \textbf{M}erging (ATM), a general framework that extends task arithmetic into an iterative process of finetuning and merging. We propose two practical use cases of ATM: (i) \textbf{\AAAA } (Privacy-Aware ATM), which enables multitask learning in settings where data cannot be shared, such as federated learning, and (ii) \textbf{\BBBB} (Post-Hoc ATM), which refines any pre-existing multitask model using only a small validation set. Furthermore, ATM allows for the integration of existing interference-resolution techniques from the model merging literature to achieve further performance gains.

To summarize, our contributions are fourfold:
\begin{enumerate}
\item We show that, under certain conditions, task vectors are equivalent to the gradients of their respective task losses.
\item We introduce Alternating Tuning and Merging (ATM), a flexible framework that generalizes task arithmetic and enables gradual multitask knowledge integration, supporting any conflict-resolution method for improved results.
\item We propose two practical applications of ATM: \AAAA, which replaces multitask learning when data sharing is restricted, and \BBBB, a post-hoc refinement applicable to any multitask model.
\item We validate ATM’s effectiveness through extensive experiments on image classification benchmarks.
\end{enumerate}


\section{Preliminary}
\paragraph{\textbf{Task Arithmetic.}}
\textit{Task arithmetic} merges models by adding their \textit{task vectors}, the parameter shifts ($\tau_i = \theta_i - \theta_0$) resulting from finetuning a base model ($\theta_0$) on different tasks $t_i$ independently. By creating a new model $\theta_{\text{merge}} = \theta_0 + \alpha \sum \tau_i$, knowledge from multiple tasks is combined into a single network, avoiding the need to store and deploy separate models.

\paragraph{\textbf{Task Vectors as Scaled Gradients.}}
Under the assumption that task-specific finetuning is performed for a single epoch using full-batch gradient descent, the task vector $\tau_i$ is exactly proportional to the negative gradient of the task loss at the pretrained parameters. Let $\mathcal{L}_{t_i}(\theta)$ denote the full-batch loss function for task $t_i$, and let $\eta$ be the learning rate. Performing one gradient step on $\theta_0$ yields: $ \theta_i = \theta_0 - \eta \nabla_\theta \mathcal{L}_{t_i}(\theta_0)$.
Thus, the task vector becomes $\tau_i = \theta_i - \theta_0 = -\eta \nabla_\theta \mathcal{L}_{t_i}(\theta_0) $, revealing that $\tau_i$ is a scaled negative gradient of the task loss evaluated at $\theta_0$. This gives a principled interpretation of task vectors as approximations of task objectives, bridging task arithmetic and optimization theory.

\section{{ATM: Alternating Merging and Tuning}}
Building upon the relationship between task vectors and gradients, we hypothesize that task arithmetic is an approximation to a single GD step over the union of all the tasks. Following this parallel, we advocate taking further update steps iteratively.

The overall framework of ATM is depicted in Fig.~\ref{fig:ATM diagram}. Specifically, we start from an initial base model $\theta_{\text{base}}^{(0)}$, we finetune it separately on each task to obtain the first-iteration task vectors $\tau_{1}^{(1)},  \dots, \tau_{|T|}^{(1)} $. These are then aggregated and added to the base model to form the next-iteration unified model $\theta_{\text{base}}^{(1)}$. 
The procedure is iterated according to the following equation:
%
%
\begin{equation}
    \theta_{\text{base}}^{(k+1)} = \theta_{\text{base}}^{(k)} + \frac{\alpha}{|T|}\sum_{t \in T}\tau_{t}^{(k)} \quad \forall k = 0, \dots, {K-1} \,.
    \label{eq:ATM}
\end{equation}
The $k$-th iteration task vector for task $t$ is obtained as $\tau^{(k)}_t = \theta_{t}^{(k)}- \theta_{\text{base}}^{(k)}$, where $\theta_{t}^{(k)}$ is a model obtained finetuning the $k$-th iteration base model $\theta_{\text{base}}^{(k)}$ on task $t$. The total number of iterations $K$ can be predefined or based on a stop condition.

In practice, each ATM iteration finetunes the current base model on all $|T|$ tasks using their respective data, producing $|T|$ task vectors that approximate the directions the model should follow to improve performance on each task.

In each iteration, ATM updates the base model by adding the mean of the current task vectors, though any conflict-resolution method from the task vector literature can be adopted. This merging step moves the model closer to the multitask optimum, while averaging ensures the update magnitude remains independent of the number of tasks. Since task vectors from previous iterations are discarded, ATM requires no more storage than standard task arithmetic, only the current base model and one task vector per task are retained.

ATM is generic and can be applied in multiple ways. For example, ATM can be used to build a multitask model from scratch using non-centralized training data and a pretrained initialization (\textbf{\AAAA}). This is tightly related to \textit{FedAVG} \citep{mcmahan2017communication} in the scenario of federated learning. Alternatively, our framework can be adopted to improve an existing merged model with a small set of validation data (\textbf{\BBBB}). The difference between the two lies in the model initialization $\theta_{\text{base}}^{(0)}$ and the data used for finetuning.

\section{Experiments}
Following \citep{task-vectors}, we use ViT-B/16 \citep{dosovitskiy2020image} as the backbone and evaluate on \textit{DTD} \citep{cimpoi14describing}, \textit{EuroSAT} \citep{helber2019eurosat}, \textit{GTSRB} \citep{Houben-IJCNN-2013}, \textit{MNIST} \citep{726791}, \textit{RESISC45} \citep{cheng2017remote}, \textit{SUN397} \citep{5539970}, and \textit{SVHN} \citep{netzer2011reading}. We compare \AAAA and \BBBB to task arithmetic (TA) \citep{task-vectors}, TIES \citep{yadavties}, breadcrumbs \citep{davari2023model}, and DARE \citep{yu2024language}. If not otherwise specified, we adhere to the author-suggested hyperparameters in the baselines whenever available. For \AAAA, we initialize the base model as the ViT-B-16 pretrained on ImageNet, whereas for \BBBB, we initialize the base model as the multitask model obtained from the application of task arithmetic. Note that while the goal of \AAAA is to build a multitask model from scratch, the goal of \BBBB is to improve upon task arithmetic starting in the same conditions as the baseline methods proposed to refine task arithmetic.

Following \citep{task-vectors}, we evaluate all methods on the held-out test sets provided by each individual dataset and build the validation set by sampling $10\%$ of the training set. We adopt classification \textit{accuracy} as the evaluation metric, which is standard in the literature.


\vspace{-1.5em}
\begin{figure}
    \centering
    \includegraphics[width=1\linewidth]{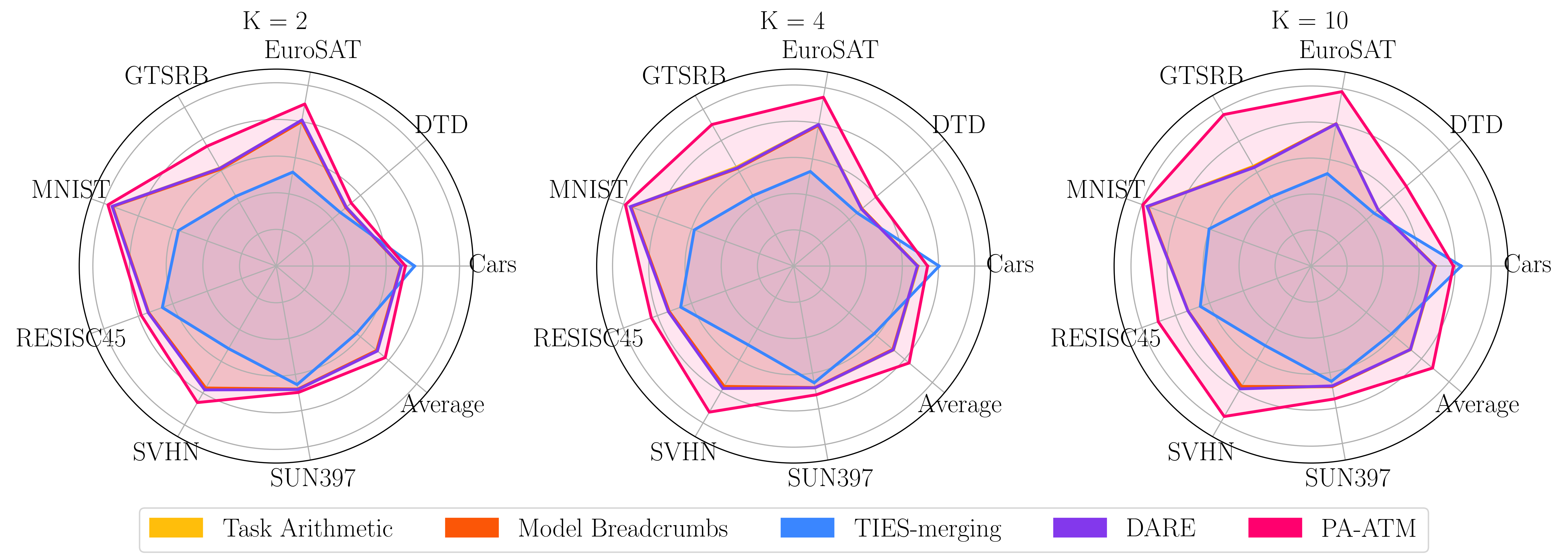}
    \caption{\AAAA vs Baselines as computational budget $K$ varies.}
    \label{fig:budget-exp-vit}
\end{figure}

\vspace{-1em}

\subsection{Impact of Epoch Distribution on Performance}

One might question whether allocating more epochs per merging step is beneficial. To show that the one-epoch convention is indeed optimal, we conduct the following experiment with \AAAA. We establish a fixed compute budget of $10$ finetuning epochs for each task, then we seek the optimal distribution of epochs among different numbers of ATM iterations. To exemplify, if $10$ epochs are distributed among $5$ iterations, then in each iteration a task is finetuned for $2$ epochs. 

\begin{figure}[ht]
    \centering
    \begin{minipage}[t]{0.45\textwidth}
        \centering
        \includegraphics[width=\linewidth]{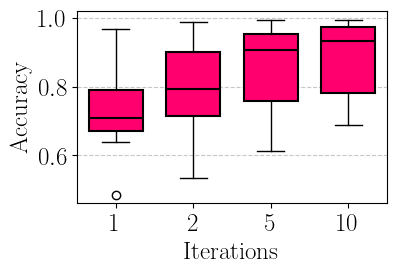}
        \caption{\AAAA multitask accuracy across budget distributions.}
        \label{fig:ATMsettings}
    \end{minipage}
    \hspace{10pt}
    \begin{minipage}[t]{0.45\textwidth}
        \centering
        \includegraphics[width=\linewidth]{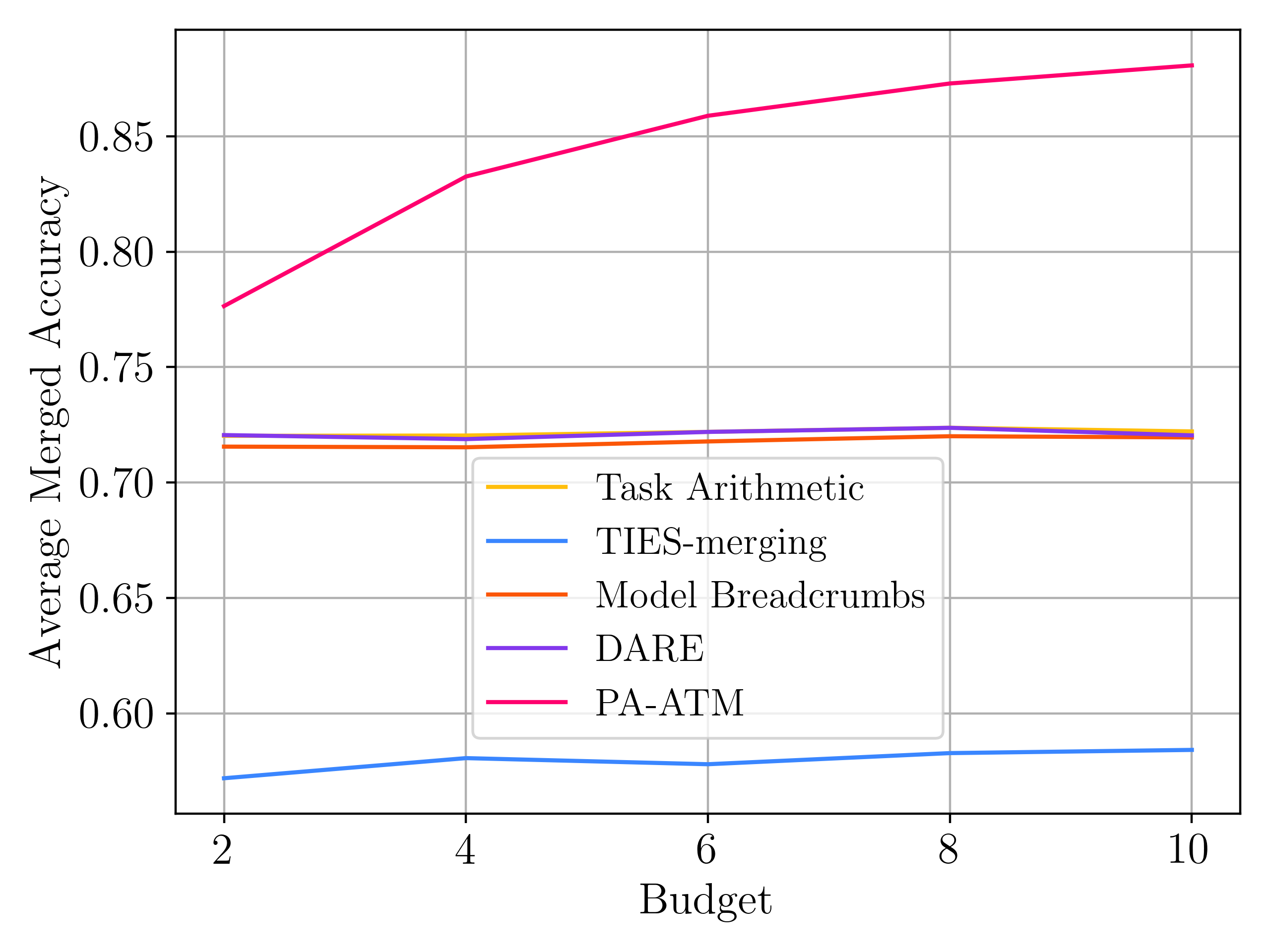}
        \caption{Average multitask accuracy as budget varies.}
        \label{fig:decreasingAccs}
    \end{minipage}
\end{figure}

As depicted in Fig.~\ref{fig:ATMsettings}, with a fixed compute budget, maximizing iterations while minimizing epochs per iteration yields the best results for ATM. This indicates that more gradual updates to the base model are preferable to abrupt ones. We observe a clear monotonous increase as more iterations are performed. Distributing $10$ epochs across $10$ iterations achieves the highest average accuracy of $88\%$, outperforming the $1$ iteration of $10$ epochs setting (analogous to task arithmetic) by around $16\%$. Full results are depicted in Appendix~\ref{atm budget distribution}.

Hence, in the rest of the paper, we adhered to the convention of merging after just a single epoch of finetuning during ATM iterations.

\subsection{Effect of Finetuning Budget}
We compare \AAAA against baselines for different budgets of finetuning epochs. We vary the total per-task finetuning epochs ($K$) within $2$, $4$, and $10$ to compare the average test accuracy across tasks. As shown in Fig.~\ref{fig:budget-exp-vit}, \AAAA consistently outperforms the baselines for all budgets. Interestingly, while \AAAA's accuracy improves with more finetuning epochs, the baseline methods remain unaffected; see Fig.~\ref{fig:decreasingAccs} and Appendix~\ref{varying budget}. In other words, \emph{more specialization of the task-specific models does not necessarily benefit the one-shot unified multitask model}. 





\subsection{ATM vs Baselines}

%

We compare the \AAAA and \BBBB to the baselines under their original data and hyperparameter setups. With an abuse of notation for the sake of readability, the epochs for \BBBB leverage validation data, unlike all other methods that use the training data. As illustrated in Table~\ref{tab:optimal settings}, \AAAA unsurprisingly and significantly outperforms the baselines. With 10 iterations of ATM, it outperforms the best-performing baseline by $15\%$ on average. If the budget is elevated to 30 iterations, the advantage grows to  $17\%$.

We emphasize again that all baselines do make use of the validation data for hyperparameter tuning, while \BBBB leverages it differently. Table~\ref{tab:optimal settings} shows that \BBBB also outperforms the baselines across all tasks but one, with an average improvement of $7\%$ over the best baseline. Thus, we postulate that estimating the task-proficiency directions through validation data, as done in \BBBB, is a better option when further finetuning is allowed.

\begin{table}
    \centering
    \resizebox{1\textwidth}{!}{
    \begin{tabular}{c cccccccc c}
        \toprule
        Method & Cars & DTD & EuroSAT  & GTSRB & MNIST & RESISC45 & SVHN & SUN397 & Average \\
        \cmidrule(lr){1-1} \cmidrule{2-9} \cmidrule(lr){10-10}
        \texttt{Pretrained} & 0.64 & 0.45 & 0.54 & 0.43 & 0.51 & 0.65 & 0.51 & 0.65 & 0.55 \\ 
        \texttt{Finetuned} & 0.87 & 0.98 & 0.98 & 0.99 & 0.99 & 0.96 & 0.97 & 0.78 & 0.94 \\ 
        \addlinespace[3pt]
        \cdashline{1-10}
        \addlinespace[3pt]
        \texttt{Task Arithmetic} & 0.69 & 0.52 & 0.83 & 0.62 & 0.95 & 0.75 & 0.75 & 0.68 & 0.73 \\
        \texttt{TIES} & 0.88 & 0.45 & 0.44 & 0.41 & 0.63 & 0.63 & 0.42 & 0.63 & 0.56 \\
        \texttt{Breadcrumbs} & 0.69 & 0.52 & 0.83 & 0.61 & 0.95 & 0.75 & 0.74 & 0.68 & 0.72 \\
        \texttt{DARE} & 0.69 & 0.52 & 0.83 & 0.61 & 0.95 & 0.75 & 0.76 & 0.68 & 0.72 \\
        \addlinespace[3pt]
        \cdashline{1-10}
        \addlinespace[3pt]
        \rowcolor{BurntOrange!15}
        \texttt{\BBBB}$_{10}$ & 0.73 & 0.61 & 0.54 & 0.95 & 0.99 & 0.89 & 0.95 & 0.71 & 0.80 \\
        \rowcolor{forestGreen!15}
        \texttt{\AAAA}$_{10}$ & 0.79 & 0.69 & 0.98 & 0.97 & 0.99 & 0.90 & 0.96 & 0.75 & 0.88 \\
        \rowcolor{forestGreen!15}
        \texttt{\AAAA}$_{30}$ & \textbf{0.82} & \textbf{0.76} & \textbf{0.99 }& \textbf{0.99} & \textbf{0.99} & \textbf{0.94} & \textbf{0.97 }& \textbf{0.76} & \textbf{0.90} \\
        \bottomrule
    \end{tabular}
    }
    \caption{Accuracy comparison under original baseline settings. \texttt{\AAAA}$_{10}$ refers to \texttt{\AAAA} employing $10$ iterations.}
    \label{tab:optimal settings}
\end{table}



\section{Related work}
\paragraph{\textbf{Model Merging.}}
Model merging combines the capabilities of independently trained models. Specifically, \citep{navon2023equivariant} explored embedding-space merging, while \citep{cycle-consistent} introduced cycle-consistent matching. For models with shared initialization, \citep{wortsman2022model} used simple averaging, while \citep{jolicoeur2023population} aligned models to the population mean for stability. Weighted averaging approaches, such as RegMean \citep{jin2022dataless} and Fisher-weighted averaging \citep{matenamerging}, optimize merging weights based on specific criteria. \citep{gradient-mismatch} linked gradient mismatch to post-merging multitask performance. Lastly, \citep{choshen2022fusing} proposed merging as an alternative to pretraining, arguing that a merged model can outperform any single finetuned checkpoint.

\paragraph{\textbf{Task Vectors.}} 
Task vectors represent parameter differences between a pretrained model and its finetuned versions \citep{task-vectors}, enabling operations like forgetting and multitask learning. Recent methods improve merging by reducing interference through sparsification \citep{panda2024lottery}, pruning \citep{yadavties, davari2023model}, masking \citep{yu2024language}, adaptive weighting \citep{yangadamerging}, or modularization \citep{yang2024representation, ortiz2024task}. Unlike these one-shot approaches, we propose an \emph{iterative} refinement strategy for enhanced multitask performance.

\paragraph{\textbf{Federated Averaging.}}
\AAAA resembles FedAVG~\citep{mcmahan2017communication} in alternating local updates and aggregation from a shared initialization. However, their underlying perspective differs. Unlike FedAVG, which targets distributed training on non-IID data, \AAAA aims at iterative model merging across tasks, bridging federated learning and model merging.

\section{Conclusions}
We connect task vectors to gradients in multitask learning, forming the foundation for our proposed Alternating Tuning and Merging (ATM) framework. ATM iteratively refines model merging, addressing the shortcomings of one-shot methods. We introduce two applications of ATM: \AAAA, which bridges model merging and federated learning, serving primarily for analysis, and \BBBB, which enhances existing multitask models using validation data only. Experiments on vision tasks show that ATM achieves strong performance with efficiency comparable to baselines. ATM is compatible with interference-resolution techniques for task vectors, and the investigation of which is left for future work.

%
%
\bibliographystyle{plainnat}
\bibliography{ref}

\section{Appendix}
\subsection{Full Results over Varying Computational Budget} \label{varying budget}
In the main paper, we pictorially illustrated the multi-task accuracy of the baselines and PA-ATM in the form of radar plots. For deeper analysis, here we report the full results of PA-ATM compared to the baselines, as the computational budget $K$ varies across 2, 4, 6, 8, and 10 epochs. 
See Table \ref{tab:vit_comparisons}.

\begin{table}[h]
    \centering
    \resizebox{\textwidth}{!}{
    \begin{tabular}{ccccccccccc}
        \toprule
        $K$ & Method & Cars & DTD & EuroSAT & GTSRB & MNIST & RESISC45 & SVHN & SUN397 & Average \\
        \midrule
        \multirow{5}{*}{2} 
        & \texttt{Task Arithmetic}  & 0.68 & 0.50 & 0.81 & 0.62 & 0.95 & 0.74 & 0.78 & 0.68 & 0.72 \\
        & \texttt{TIES}             & 0.75 & 0.46 & 0.52 & 0.44 & 0.57 & 0.66 & 0.52 & 0.66 & 0.57 \\
        & \texttt{Breadcrumbs}      & 0.68 & 0.50 & 0.80 & 0.61 & 0.95 & 0.74 & 0.77 & 0.68 & 0.72 \\ 
        & \texttt{DARE}             & 0.68 & 0.50 & 0.81 & 0.62 & 0.95 & 0.74 & 0.78 & 0.68 & 0.72 \\
        & \texttt{PA-ATM}           & 0.71 & 0.53 & 0.90 & 0.75 & 0.98 & 0.78 & 0.86 & 0.70 & 0.78 \\
        \midrule
        \multirow{5}{*}{4} 
        & \texttt{Task Arithmetic}  & 0.69 & 0.49 & 0.79 & 0.63 & 0.96 & 0.74 & 0.78 & 0.68 & 0.72 \\
        & \texttt{TIES}             & 0.80 & 0.46 & 0.53 & 0.45 & 0.58 & 0.66 & 0.50 & 0.65 & 0.58 \\
        & \texttt{DARE}             & 0.68 & 0.49 & 0.79 & 0.63 & 0.96 & 0.74 & 0.78 & 0.68 & 0.72 \\
        & \texttt{Breadcrumbs}      & 0.68 & 0.49 & 0.79 & 0.62 & 0.96 & 0.73 & 0.77 & 0.68 & 0.76 \\ 
        & \texttt{PA-ATM}           & 0.74 & 0.59 & 0.95 & 0.90 & 0.99 & 0.84 & 0.93 & 0.72 & 0.83 \\
        \midrule
        \multirow{5}{*}{6} 
        & \texttt{Task Arithmetic}  & 0.69 & 0.49 & 0.80 & 0.64 & 0.96 & 0.74 & 0.79 & 0.68 & 0.72 \\
        & \texttt{TIES}             & 0.82 & 0.45 & 0.50 & 0.45 & 0.57 & 0.66 & 0.51 & 0.65 & 0.58 \\
        & \texttt{DARE}             & 0.69 & 0.49 & 0.80 & 0.63 & 0.96 & 0.74 & 0.79 & 0.68 & 0.72 \\
        & \texttt{Breadcrumbs}      & 0.69 & 0.48 & 0.79 & 0.63 & 0.96 & 0.73 & 0.77 & 0.68 & 0.72 \\
        & \texttt{PA-ATM}           & 0.77 & 0.64 & 0.97 & 0.95 & 0.99 & 0.87 & 0.95 & 0.74 & 0.86 \\
        \midrule
        \multirow{5}{*}{8} 
        & \texttt{Task Arithmetic}  & 0.70 & 0.49 & 0.79 & 0.65 & 0.96 & 0.73 & 0.78 & 0.68 & 0.72 \\
        & \texttt{TIES}             & 0.83 & 0.46 & 0.52 & 0.45 & 0.60 & 0.66 & 0.49 & 0.65 & 0.58 \\
        & \texttt{DARE}             & 0.69 & 0.49 & 0.79 & 0.64 & 0.96 & 0.73 & 0.79 & 0.68 & 0.72 \\
        & \texttt{Breadcrumbs}      & 0.69 & 0.49 & 0.79 & 0.64 & 0.96 & 0.73 & 0.77 & 0.68 & 0.72 \\
        & \texttt{PA-ATM}           & 0.78 & 0.67 & 0.98 & 0.96 & 0.99 & 0.89 & 0.96 & 0.74 & 0.87 \\
        \midrule
        \multirow{5}{*}{10} 
        & \texttt{Task Arithmetic}  & 0.69 & 0.49 & 0.80 & 0.64 & 0.97 & 0.73 & 0.79 & 0.68 & 0.72 \\
        & \texttt{TIES}             & 0.83 & 0.46 & 0.52 & 0.44 & 0.60 & 0.65 & 0.51 & 0.65 & 0.58 \\
        & \texttt{DARE}             & 0.68 & 0.49 & 0.80 & 0.63 & 0.97 & 0.73 & 0.79 & 0.68 & 0.72 \\
        & \texttt{Breadcrumbs}      & 0.69 & 0.49 & 0.80 & 0.64 & 0.97 & 0.73 & 0.77 & 0.68 & 0.72 \\
        & \texttt{PA-ATM}           & 0.79 & 0.69 & 0.98 & 0.97 & 0.99 & 0.90 & 0.96 & 0.75 & 0.88 \\
        \bottomrule
    \end{tabular}
    }
    \caption{PA-ATM vs. Baselines as budget varies (\textit{ViT-B-16})}
    \label{tab:vit_comparisons}
\end{table}

\subsection{Full Results of Varying ATM Budget Distribution} \label{atm budget distribution}
We show the full accuracy results as we vary the distribution of 10 epochs into different numbers of PA-ATM iterations. See Table \ref{tab:atm_distribution}.
\begin{table}[h]
    \centering
    \resizebox{\textwidth}{!}{
    \begin{tabular}{ccccccccccc}
        \toprule
        Iterations & Epochs/iteration & Cars & DTD & EuroSAT & GTSRB & MNIST & RESISC45 & SVHN & SUN397 & Average \\
        \midrule
        \multirow{1}{*}{1} 
        & \texttt{10}  & 0.6886 & 0.4858 & 0.8027 & 0.6402 & 0.9681 & 0.7276 & 0.7850 & 0.6792 & 0.7222 \\
        \midrule
        \multirow{1}{*}{2} 
        & \texttt{5}  & 0.7203 & 0.5337 & 0.9093 & 0.8004 & 0.9862 & 0.7857 & 0.8971 & 0.6964 & 0.7911 \\
        \midrule
        \multirow{1}{*}{5} 
        & \texttt{2}  & 0.7668 & 0.6126 & 0.9616 & 0.9395 & 0.9921 & 0.8716 & 0.9494 & 0.7279 & 0.8527 \\
        \midrule
        \multirow{1}{*}{10} 
        & \texttt{1}  & 0.7904 & 0.6897 & 0.9837 & 0.9713 & 0.9943 & 0.9032 & 0.9644 & 0.7490 & 0.8808 \\
        \bottomrule
    \end{tabular}
    }
    \caption{PA-ATM accuracies for different budget distributions}
    \label{tab:atm_distribution}
\end{table}

\end{document}